\documentclass[review,numbers]{elsarticle} 
\usepackage{amsmath,amssymb,amsfonts}
\usepackage{graphicx,xcolor}
\usepackage{changepage}
\usepackage{textcomp}
\usepackage{tabularx}
\usepackage{booktabs}
\usepackage{subfig}
\usepackage{lineno}
\usepackage{pdflscape}
\makeatletter
\def\ps@pprintTitle{%
  \let\@oddhead\@empty
  \let\@evenhead\@empty
  \def\@oddfoot{\reset@font\hfil\thepage\hfil}
  \let\@evenfoot\@oddfoot
}
\makeatother

\begin{document}
\begin{frontmatter}
\title{A multi-platform LiDAR dataset for standardized forest inventory measurement at long term ecological monitoring sites}

\newcommand{\orcidauthorA}{0009-0005-1892-7773} %
\newcommand{\orcidauthorB}{0000-0003-3794-5510} %
\newcommand{\orcidauthorC}{0000-0001-6546-6459} %
\newcommand{\orcidauthorD}{0000-0003-2675-9421} %
\newcommand{\orcidauthorE}{0000-0001-7094-4582} %

\author[eng-unibz]{Michael R. Chang} \ead{mchang@unibz.it}
\author[sci-unibz]{Anna Candotti} \ead{anna.candotti@unibz.it}
\author[eng-unibz]{Karl von Ellenrieder}
\ead{karl.vonellenrieder@unibz.it}
\author[sci-unibz,for-unibz]{Enrico Tomelleri}
\ead{etomelleri@unibz.it}
\author[unitn,eng-unibz]{Marco Camurri}
\ead{marco.camurri@unitn.it}

\affiliation[eng-unibz]{organization={Faculty of Engineering, Free University of Bozen-Bolzano},
             addressline={via Bruno Buozzi 1},
             city={Bolzano},
             postcode={39100},
             state={BZ},
             country={Italy}}
\affiliation[unitn]{organization={Department of Industrial Engineering, University of Trento},
             addressline={via Sommarive 9},
             city={Trento},
             postcode={38123},
             state={TN},
             country={Italy}}
\affiliation[sci-unibz]{organization={Faculty of Agricultural, Environmental and Food Sciences, Free University of Bozen-Bolzano},
             addressline={piazza Università, 5},
             city={Bolzano},
             postcode={39100},
             state={BZ},
             country={Italy}}
\affiliation[for-unibz]{organization={Competence Centre for Mountain Innovation Ecosystems, Free University of Bozen-Bolzano}}

\begin{abstract}
We present a curated multi-platform LiDAR reference dataset from an instrumented ICOS forest plot, explicitly designed to support calibration, benchmarking, and integration of 3D structural data with ecological observations and standard allometric models. The dataset integrates UAV-borne laser scanning (ULS) to measure canopy coverage, terrestrial laser scanning (TLS) for detailed stem mapping, and backpack mobile laser scanning (MLS) with real-time SLAM for efficient sub-canopy acquisition. We focus on the control plot with the most complete and internally consistent registration, where TLS point clouds ($\sim$333 million points) are complemented by ULS and MLS data capturing canopy and understory strata. Marker-free, SLAM-aware protocols were used to reduce field and processing time, while manual and automated methods were combined. Final products are available in LAZ and E57 formats with UTM coordinates, together with registration reports for reproducibility. The dataset provides a benchmark for testing registration methods, evaluating scanning efficiency, and linking point clouds with segmentation, quantitative structure models, and allometric biomass estimation. By situating the acquisitions at a long-term ICOS site, it is explicitly linked to 3D structure with decades of ecological and flux measurements. More broadly, it illustrates how TLS, MLS, and ULS can be combined for repeated inventories and digital twins of forest ecosystems.

\end{abstract}


\end{frontmatter}

\section{Summary}
Ecosystem monitoring networks such as the Integrated Carbon Observation System (ICOS)~\cite{gielen2017integrated} rely on repeated inventories to link matter and energy fluxes with above-ground biomass (AGB). Current protocols are based on tree diameter, height, and allometric models, which provide continuity but also propagate uncertainties due to model selection and limited calibration data~\cite{demol2022}. Terrestrial laser scanning (TLS) has been tested as a non-destructive alternative, reconstructing tree volume and AGB through quantitative structure models (QSMs)~\cite{holvoet2025}. While TLS reduces bias compared to allometric scaling, especially for large trees, it remains time-intensive and vulnerable to occlusion, limiting its use for repeated inventories~\cite{boucher2021sampling}.

Emerging LiDAR modalities offer complementary advantages. UAV-borne laser scanning (ULS) captures canopy and terrain structure efficiently, whereas mobile laser scanning (MLS) enables rapid sub-canopy acquisition~\cite{muhojoki_benchmarking_2024, balestra_lidar_2024}. Their performance under alpine conditions—where steep slopes and dense understory increase registration uncertainty~\cite{ritter_towards_2020}—has yet to be systematically assessed, and long-term strategies for standardization across monitoring sites remain under development~\cite{murtiyoso_virtual_2023}. Recent advances such as VIO-enabled TLS~\cite{metzler2024vio} and SLAM-based MLS~\cite{chen2020sloam} further motivate testing of marker-free registration workflows in operational inventories.

\begin{figure}
\centering
\subfloat[Alignment of all three modalities: ALS, TLS, MLS.] %
{\includegraphics[width=9.0cm]{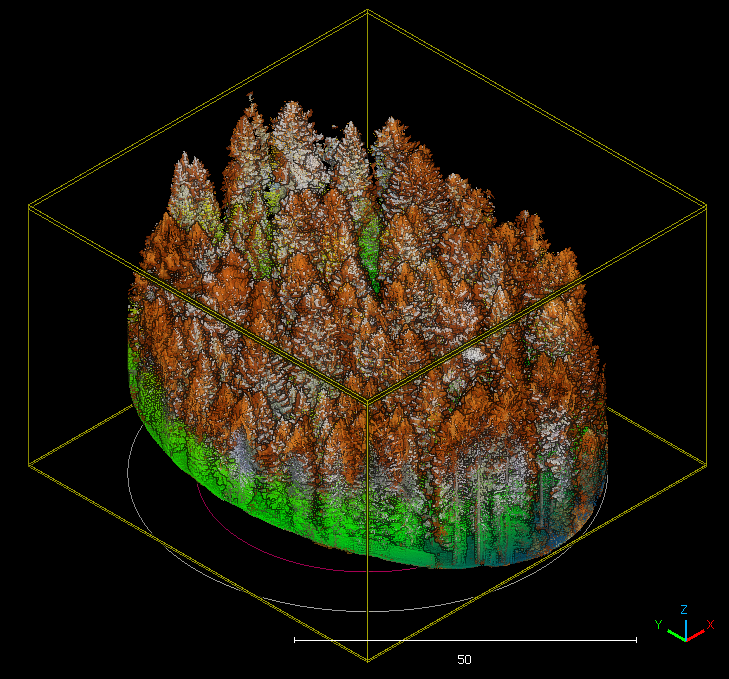}}
\hfill
\subfloat[Cloud-to-cloud distance: MLS to ALS]{\includegraphics[width=9.0cm]{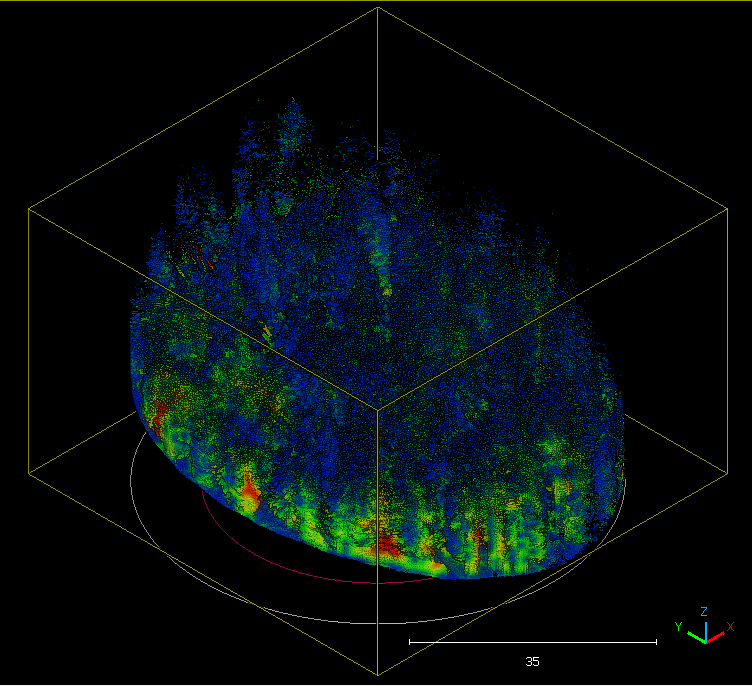}}
\caption{Multi-platform LiDAR integration and registration validation for CP2 forest inventory site. Panel~(a) demonstrates complementary coverage characteristics of aerial (ALS, orange), mobile (MLS, blue-green), and terrestrial (TLS, white) laser scanning modalities. Panel (b) Cross-platform registration quality assessment showing cloud-to-cloud distance analysis between MLS and ALS datasets. Color mapping indicates geometric alignment accuracy with blue regions showing excellent correspondence (<2cm), green showing good alignment (2-5cm), and warmer colors indicating areas where there simply were no ALS points. This is one example of many pairwise examinations possible with this dataset.}
\label{fig:multiplatform_integration}
\end{figure}

The dataset described in this paper and available in \cite{dataset} addresses this challenge through a case study at the ICOS Class~2 site Renon (IT-Ren) in the Italian Alps, a subalpine Norway spruce (\textit{Picea abies}) forest characterized by steep terrain and complex canopy structure. It integrates TLS, MLS, and ULS acquisitions into co-registered point clouds aligned with quality goals outlined in the RINGO project~\cite{ringo} (uniform point density, accurate registration, minimized occlusion). Platform-specific coverage of canopy, trunk, and understory strata is evident in the vertical point-density distributions (Figure~\ref{fig:height_distribution_comparison}), and the accuracy of cross-platform alignment is supported by cloud-to-cloud comparisons (Figure~\ref{fig:multiplatform_integration}).

This dataset demonstrates how multi-platform acquisitions can support forest inventories across ICOS sites. It provides a reusable reference for benchmarking registration methods~\cite{castorena_forestalign_2024}, evaluating SLAM- and VIO-based workflows~\cite{boche2024tightlycoupledlidarvisualinertialslamlargescale, oh2024evaluation}, refining biomass estimation through segmentation, QSM, and allometry~\cite{puliti_benchmarking_2024, wielgosz_segmentanytree_2024}, and assessing the efficiency of alternative scanning strategies. Beyond this single site, the integration principles shown here can be applied across the ICOS network and other monitoring programs, including operational contexts such as national forest inventories~\cite{kuekenbrink2025} and emerging autonomous deployments~\cite{mattamala2024autonomous, cheng2024treescope, border2024osprey}. More broadly, the dataset contributes to the development of digital twins of forest ecosystems, providing opportunities to test when virtual 3D reconstructions become sufficiently accurate to complement or partially replace traditional inventories, and to link digital forest representations with long-term ecological and flux records.

The dataset focuses on a single ICOS control plot selected for its comprehensive ecological instrumentation and optimal cross-platform LiDAR registration quality. This deliberate focus enables high-density, multi-platform acquisitions that are impractical at scale, but essential for method development, protocol standardization, and calibration.

Rather than serving as a statistical sample of forest conditions, this dataset is intended as a reference and benchmarking resource, enabling users to (i) evaluate multi-platform LiDAR registration and sampling strategies, (ii) test forest inventory algorithms against a well-instrumented ecological plot, and (iii) link 3D forest structure to independent ICOS ecological observations using established allometric relationships.

\section{Data Description}
In the following sections we describe the experimental site and design, the different types of laser scanning data used in this dataset, and how they were integrated.

\subsection{Site Description and Experimental Design}
This dataset presents multi-platform LiDAR acquisitions with integrated aerial, terrestrial, and mobile laser scanning from the Italian Alps' Renon ICOS Class~2 station (IT-Ren, [46.58686, 11.43369]). The site exemplifies challenging subalpine conditions at 1735~m elevation, where its uneven-aged, ~200-year-old Norway spruce (\emph{Picea abies}) canopy creates complex scanning scenarios across steep terrain and dense understory. Long-term ecological and flux measurements have been maintained at IT-Ren since 1997~\cite{icos2024renon}, making it an ideal reference site for testing novel LiDAR-based forest inventory approaches.

Following ICOS sampling protocols~\cite{ringo, gielen2017integrated}, data were collected across three 25~m-radius control plots (CP1–3). CP2 is the focus of this work due to optimal cross-platform registration quality and comprehensive sensor coverage (see Figure~\ref{fig:multiplatform_integration}). CP2 also hosts additional ecological instrumentation, including sensors for litterfall, soil moisture, soil temperature, and soil water potential, allowing integration of LiDAR data with ancillary ecosystem observations. These data provide an exemplary TLS registration for non-destructive above-ground biomass (AGB) estimation~\cite{demol2022, holvoet2025}, with accompanying methodology to replicate the full pipeline from field collection to a reproducible ``digital forest'' product.

The dataset includes three primary scanning modalities with distinct conventions. Aerial data technically qualify as ULS (Unmanned Laser Scanning) collected via UAV rather than traditional aircraft-based ALS, providing finer spatial resolution from lower altitude~\cite{naesset1997estimating, naesset2002predicting}. MLS data were collected using an oblique-mounted backpack payload, an important distinction in forestry where sensor height influences understory occlusion and automatic registration accuracy~\cite{su2021backpack, bauwens2016forest}. TLS data were acquired with a Leica BLK360 using real-time visual–inertial odometry (VIO) for efficient point cloud co-registration \emph{in situ}, an approach more typical of MLS systems but still novel for static TLS instruments~\cite{metzler2024vio}. This combination of modalities enables evaluation of platform-specific strengths and weaknesses—ALS canopy bias, TLS occlusion effects~\cite{boucher2021sampling}, and MLS trajectory constraints—within a standardized ICOS monitoring framework.

\subsection{Multi-Platform Integration: Inter-modal registration objects}

Cross-platform integration is the central contribution of this dataset, bringing together aerial, terrestrial, and mobile LiDAR acquisitions within a standardized ICOS framework. By combining UAV-based canopy coverage, TLS stem detail, and MLS sub-canopy trajectories, the dataset captures complementary perspectives on forest structure. Platform-specific characteristics are summarized in Table~\ref{tab:platform_specs}, while differences in vertical sampling are shown in Figure~\ref{fig:height_slice_comparison}. 

Despite order-of-magnitude differences in point densities, ground surfaces derived from cloth simulation filtering (CSF) were consistent across modalities, yielding comparable digital terrain models (Figure~\ref{fig:ground_dtm_comparison}). This consistency underscores the robustness of the integration and provides confidence that the co-registered point clouds can support standardized height normalization and cross-platform analysis in complex structural conditions.

\begin{table}
\caption{Technical specifications and operational characteristics of three LiDAR platforms used in forest inventory\label{tab:platform_specs}}
{%
\begin{tabularx}{\textwidth}{XXXX}
\toprule
\textbf{Modality} & \textbf{ALS/ULS} & \textbf{TLS} & \textbf{MLS} \\
\midrule
\textbf{Payload} & \textbf{Yellowscan Mapper+} & \textbf{Leica BLK360} & \textbf{Frontier ORI-DRS} \\
\textbf{LiDAR Make \& Model} & Livox AVIA & Proprietary & HESAI XT32 \\
\textbf{Accuracy \& Range} & 35-40~mm, 100~m & 4-6~mm, 25~m & 10-20~mm, 25~m \\
\textbf{Scan Rate (pts/sec)} & Up to 240,000 & 360,000 & 640,000 (@ 10 Hz) \\
\hline
\textbf{GNSS Integration} & GNSS: 1 cm PPK & n/a (External: GCP targets) & SLAM-dependent \\
\textbf{Camera Array} & N/A (lacks optional RGB) & 4-camera, 13MP each, 150MP spherical & 3-camera, AlphaSense fisheye \\
\textbf{IMU Integration} & GNSS-IMU fusion & VIO localization & LIO, multi-sensor fusion \\
\textbf{Registration Method} & PPK georeferencing & Manual + VIO prior & SLAM + loop closure \\
\hline
\textbf{Under-canopy Access} & Limited (canopy penetration only) & Complete (360° coverage) & Good (mobile ground access) \\
\textbf{Collection Speed} & Fast (hectares/hour) & Slow (hours/plot, 5-10 min/setup) & Moderate (\~1 plot/hour) \\
\bottomrule
\end{tabularx}
}
\label{tab:lidar-tech-specs}
\end{table}

\begin{table}
\caption{Ground and vegetation stratification metrics for CP2 scanning platforms\label{tab:stratification_metrics}}
\newcolumntype{L}{>{\centering\small\arraybackslash}l}

\begin{tabularx}{\columnwidth}{p{4cm}*{6}{L}}
\toprule
\textbf{Metric} & \multicolumn{2}{l}{\textbf{ULS}} & \multicolumn{2}{l}{\textbf{TLS}} & \multicolumn{2}{l}{\textbf{MLS}} \\
\cmidrule(lr){2-3} \cmidrule(lr){4-5} \cmidrule(lr){6-7}
& \textbf{Count} & \textbf{\%} & \textbf{Count} & \textbf{\%} & \textbf{Count} & \textbf{\%} \\
\midrule
\textbf{Total Points (35~m radius)} & 6,851,016 & 100.0 & 203,657,957 & 100.0 & 32,859,532 & 100.0 \\
\textbf{Ground Points (CSF)} & 532,888 & 13.6 & 54,148,663 & 26.6 & 8,761,875 & 26.7 \\
\textbf{Vegetation Points} & 5,915,620 & 86.4 & 149,589,793 & 73.4 & 24,145,090 & 73.3 \\
\midrule
\multicolumn{7}{l}{\textit{Height Stratification Results}} \\
\textbf{Understory (0-1.4~m)} & 47,944 & 0.8 & 12,257,657 & 8.2 & 5,198,507 & 21.5 \\
\textbf{Trunks (1.4-5~m)} & 37,962 & 0.6 & 53,311,535 & 35.6 & 15,757,716 & 65.3 \\
\textbf{Midcanopy (5-13~m)} & 1,482,391 & 25.1 & 63,582,670 & 42.5 & 3,031,980 & 12.6 \\
\textbf{Canopy Tops ($>$13~m)} & 4,011,323 & 67.8 & 20,437,931 & 13.7 & 156,887 & 0.6 \\
\midrule
\textbf{Height Range (Min-Max)} & \multicolumn{2}{c}{-1.12 to 35.26~m} & \multicolumn{2}{c}{-1.15 to 33.54~m} & \multicolumn{2}{c}{-1.28 to 32.87~m} \\
\bottomrule
\end{tabularx}

\end{table}

\subsection{Aerial/Unmanned Laser Scanning Data}

Data outputs were processed using YellowScan proprietary software with post-process kinematic (PPK) GNSS georeferencing, achieving approximately 1 cm global positioning accuracy and point density of 1,344 points per square meter across the 13.5-hectare study area with an output 5.07 GB LAS file containing 181.72 million points.

Raw data are provided in LAS 1.2 format. Ground point classification was performed using Cloth Simulation Filter algorithms, with manual noise removal and geometric clipping to the CP2 study area performed in CloudCompare.  All coordinates are maintained in UTM Zone 32N reference system (EPSG:32632), serving as the geometric foundation for multi-platform registration with terrestrial and mobile laser scanning datasets.

\subsection{Terrestrial Laser Scanning Data}

Terrestrial data were collected using a Leica BLK360 imaging scanner configured for target-free registration across 54 individual scan positions distributed throughout the CP2 study area. Data outputs comprise and RGB-colorized registered network cloud of 333.5 million points after bundle optimization and 1 cm voxel downsampling. Point densities exceed 2000 points/m² in optimal viewing areas. %

\begin{figure}
\centering
\includegraphics[width=0.9\columnwidth]{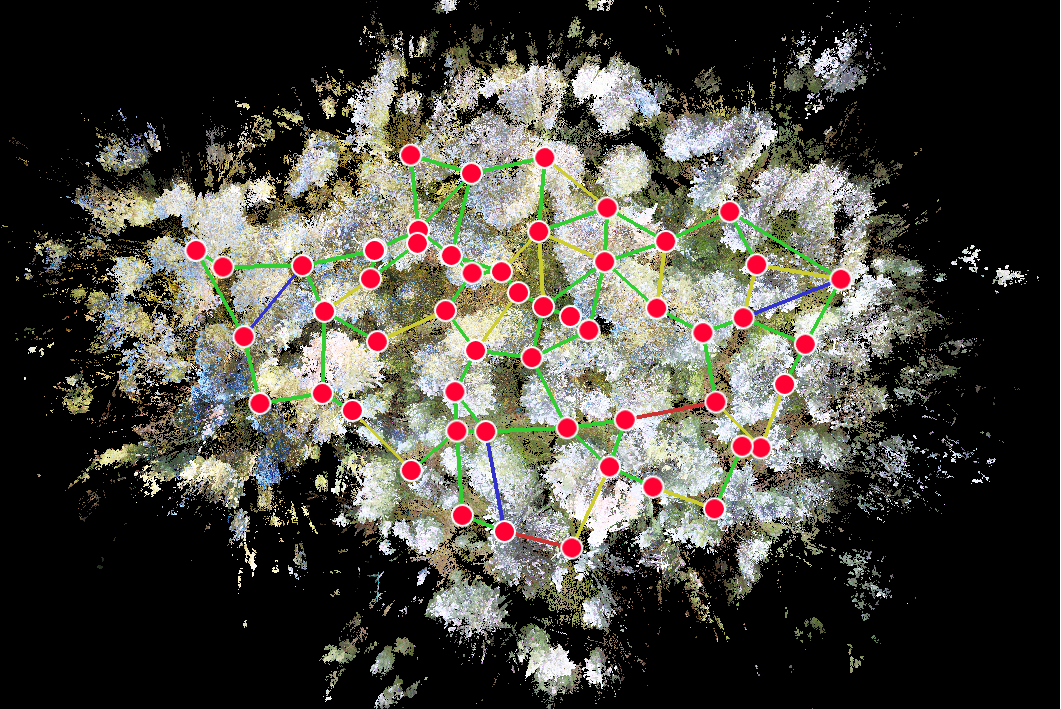}
\caption{TLS scan network topology and registration quality assessment for CP2. Link colors indicate registration accuracy: green ($\leq$ 0.018~m), yellow ($\leq$ 0.030~m), red ($>$ 0.030~m), and blue (manually aligned links). Top-down view of cloud in Register360 arbitrary reference frame.}
\label{fig:tls_network}
\end{figure}

Network registration is achieved through 85 inter-scan links with global bundle error of 5.0~mm. Individual link registration errors range from 0.0~mm to 20.0~mm with mean alignment accuracy of 14.0~mm. Setup-to-setup overlap varies from 12\% to 65\% with network average of 24\%, reflecting adaptive positioning based on local geometric constraints and sight line availability.

\begin{table}
\centering
\caption{TLS Network Registration Summary}
\begin{tabular}{lr}
\toprule
\textbf{Network Parameter} & \textbf{Value} \\
\midrule
Setup Count & 54 \\
Link Count & 85 \\
Total Points (registered) & 333.5 million \\
Global Bundle Error & 5.0~mm \\
Mean Link Error & 14.0~mm \\
Error Range & 0.0–20.0~mm \\
Average Overlap & 24\% \\
\bottomrule
\end{tabular}
\label{tab:register360-summary}
\end{table}

Raw data are archived as individual E57 format files preserving complete sensor metadata and registration network topology for alternative processing workflows. Registered output is provided in LAZ format with UTM Zone 32N coordinate system (EPSG:32632) for multi-platform integration. Processing documentation includes complete Leica Register360 bundle adjustment report detailing pairwise link quality metrics summarized in Table \ref{tab:register360-summary}.

\subsection{Mobile Laser Scanning Data}

Mobile data were collected using a backpack-mounted LiDAR system (Frontier device, University of Oxford) employing real-time SLAM processing with loop closure optimization over approximately 58 minutes of field data collection. Data outputs comprise multiple processing levels organized within structured directories containing both processed point cloud products and raw sensor data archives.

\begin{figure}
\centering
\includegraphics[width=\columnwidth]{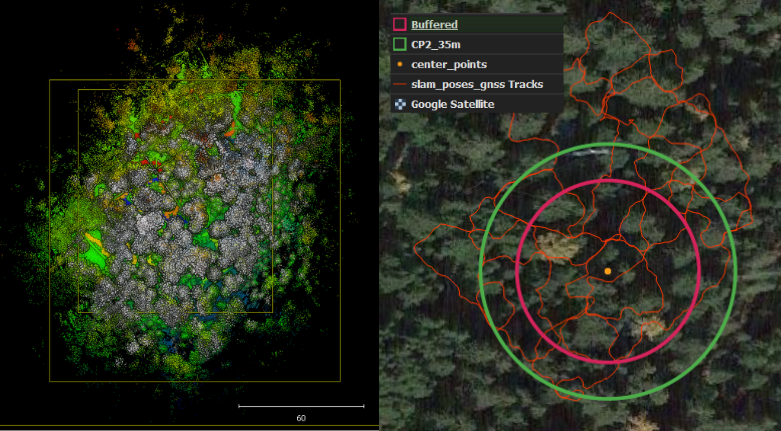}
\caption{Site context showing CP2 location within the broader ICOS Renon forest monitoring
station. Left panel displays multi-platform point cloud data visualization in CloudCompare software with the TLS (terrestrial) and the MLS (mobile) datasets co-registered. Right panel shows QGIS-based site overview with CP2 control plot boundaries (red and green circles indicate radii of 25~m and 35~m, respectively), SLAM trajectory paths (orange lines), center point markers (yellow dot), and satellite imagery context. The buffered analysis zones enable edge-effect evaluation and multi-scale forest inventory assessment within the ICOS standardized monitoring framework.}
\label{fig:site_context}
\end{figure}

Primary data products include combined point cloud datasets available in both full-resolution and downsampled formats, comprehensive SLAM pose trajectories provided in multiple coordinate systems and file formats (CSV, PCD~\cite{pcdformat}, KML, g2o~\cite{g2o}), and raw GNSS measurements enabling independent georeferencing validation. Individual point cloud segments preserve temporal sequencing through timestamp-based naming conventions, supporting trajectory reconstruction and quality assessment workflows.

Raw sensor data are archived as 14 sequential rosbag~\cite{rosbag} files spanning the complete data collection session (58 minutes), each approximately 10~GB in size containing synchronized LiDAR point streams, IMU measurements, GNSS observations, and system telemetry. SLAM trajectory accuracy and reconstruction is verified by the global registration correspondence to both TLS (Figure~\ref{fig:site_context}, left) and ALS (Figure~\ref{fig:multiplatform_integration}, right). Processing outputs are organized to support both inter-platform analysis (combined\_cloud.pcd is transformed and clipped to produce final UTM frame LAS outputs), and intra-platform (individual\_clouds/ and slam\_poses) alignment.

Data organization includes individual timestamped point cloud files corresponding to discrete SLAM keyframes, integrated pose solutions in both local SLAM coordinates and globally registered UTM Zone 32N reference system (EPSG:32632), and comprehensive sensor data preservation through rosbag archives enabling complete processing workflow reconstruction and alternative algorithm development.

\subsection{Data Formats and File Structure}

The dataset employs a hierarchical directory structure organizing data from raw sensor measurements through platform-specific processing to final integrated products. The file structure is shown in Table~\ref{tab:file_structure}. Top-level files provide immediate access to co-registered study area datasets, while platform-specific directories preserve complete processing workflows and validation documentation.

\begin{landscape}
\begin{table}[h!]
\centering
\caption{Multi-Platform Dataset Hierarchical Structure\label{tab:file_structure}}
\footnotesize
\begin{tabularx}{\linewidth}{p{0.28\linewidth}p{0.38\linewidth}p{0.08\linewidth}p{0.12\linewidth}}
\toprule
\textbf{Directory Level} & \textbf{Data Product} & \textbf{Format} & \textbf{Size (GB)} \\
\midrule
\multicolumn{4}{l}{\textit{Top-Level Integrated Products}} \\
& \texttt{renon\_CP2\_blk360\_UTM\_clip\_35m.las} & LAS & 7.2 \\
& \texttt{renon\_CP2\_blk360\_UTM\_clip\_35m.laz} & LAZ & 1.8 \\
& \texttt{renon\_CP2\_Frontier\_UTM\_clip\_35m.las} & LAS & 1.2 \\
& \texttt{Renon\_CP2\_Mapper\_clip\_35m.las} & LAS & 0.2 \\
\midrule
\multicolumn{4}{l}{\textit{Platform-Specific Directories}} \\
\texttt{YS\_Mapper+/} & & & \\
& \texttt{Renon\_Mapper.las} & LAS & 5.3 \\
& \texttt{Tmat4x4\_CP2\_blk360\_to\_Mapper\_UTM.txt} & TXT & $<$0.1 \\
\texttt{2024-11-13-leicaBLK360-icos-CP2/} & & & \\
& \texttt{renon\_CP2\_blk360\_combined.las} & LAS & 6.5 \\
& \texttt{E57\_setup\_archive/} (54 individual scans) & E57 & 8.5 \\
& \texttt{20241113blk360icosCP2.pdf} & PDF & $<$0.1 \\
\texttt{2024-11-21-frontier-icos-CP2/} & & & \\
& \texttt{combined\_cloud.pcd} & PCD & 2.8 \\
& \texttt{individual\_clouds/} (1160+ point clouds) & PCD & 12.0 \\
& \texttt{slam\_poses\_gnss.csv} & CSV & $<$0.1 \\
& \texttt{slam\_pose\_graph.g2o} & G2O & $<$0.1 \\
\midrule
\multicolumn{4}{l}{\textit{Raw Sensor Data}} \\
\texttt{2024-11-21-frontier-icos-CP2/bags/} & & & \\
& \texttt{frontier\_*\_*.bag} (14 sequential files) & ROS Bag & 140.0 \\
& \texttt{raw\_gnss\_measurements\_enu.csv} & CSV & $<$0.1 \\
\bottomrule
\end{tabularx}
\end{table}
\end{landscape}

The hierarchical organization of data lineage from raw sensor measurements to final products draws parallel between scan modalities. Top-level files provide immediate access to co-registered study area datasets suitable for comparative analysis. Platform-specific directories contain \textbf{complete, internally-consistent processing outputs} including individual scan archives that support algorithm development and validation studies. To facilitate access, the platform specific directories are provided directly as ZIP files. Raw sensor data preservation through rosbag archives \cite{rosbag} enables complete workflow reconstruction and alternative processing approaches. Transformation matrices and registration reports document geometric relationships between coordinate systems and provide validation metrics for multi-platform integration quality assessment.

\section{Methods}

\subsection{Field Data Collection Protocols}
Data collection was conducted between July-November 2024 at the Renon ICOS station under optimal weather conditions (dry, minimal wind, no precipitation) to ensure consistent data quality across all platforms. Collection proceeded sequentially: aerial surveys were first conducted to establish georeferenced baseline, followed by terrestrial networks for detailed structural mapping, and mobile surveys for comprehensive understory coverage.

\subsubsection{Aerial/Unmanned Laser Scanning Protocol}

ALS data were collected using a DJI Matrice 300 equipped with a YellowScan Mapper+ LiDAR payload. Data processing was performed using YellowScan Mapper software with post-process kinematic (PPK) georeferencing. 

The flight path followed a systematic grid pattern at 60-70~m above ground level. Minimal human intervention was required for alignment and georeferencing of the ALS data. The point cloud was filtered and using YellowScan software. 

\subsubsection{Terrestrial Laser Scanning Protocol}

Unlike previous approaches that rely on physical registration targets, our TLS data were collected using a Leica BLK360 scanner featuring integrated visual-inertial odometry (VIO) for improved setup positioning estimation. This technological advancement helps bridge the gap between traditional static TLS and dynamic MLS approaches, reducing field time while maintaining data quality.

\begin{figure}
\centering
\includegraphics[width=\textwidth]{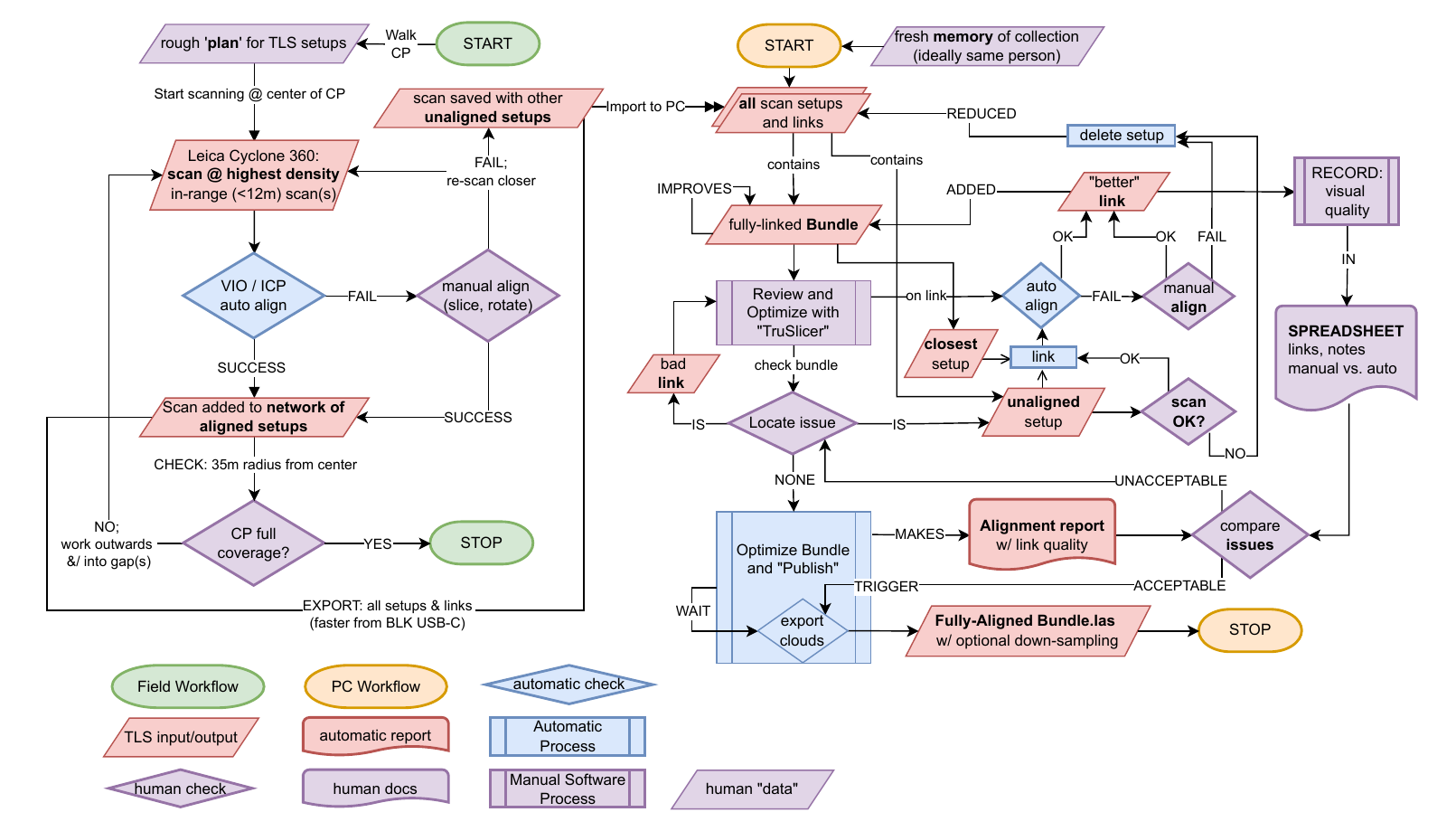}
\caption{Complete TLS collection and processing workflow integrating field and desktop protocols. The field workflow (left) emphasizes adaptive scan positioning with real-time VIO/ICP alignment quality assessment, while the PC workflow (right) details bundle optimization and link quality validation. Decision points ensure coverage completeness within a 35~m radius while maintaining registration quality through manual intervention when automatic methods fail. The integrated approach reduces field time through in-situ quality control and enables efficient post-processing through documented link relationships.}
\label{fig:tls_protocol}
\end{figure}

The BLK360's shorter effective range (25~m) compared to longer-range systems enables more detailed scan networks while field software facilitates setup optimization around tree occlusions. The Visual-Inertial Odometry integration provides real-time positioning feedback that enables adaptive collection patterns responsive to forest density variations.

The scanning process followed an adaptive network-based protocol (Figure~\ref{fig:tls_protocol}), starting from the plot center and expanding outward based on real-time registration quality assessment. Point clouds were collected and aligned using in-field manual registration through the Leica Cyclone 360 field collection interface application. This process was expanded based on successful or unsuccessful automatic and manual registrations until the full plot radius (initially 25~m, later expanded to 30~m) was covered.

For consistency with MLS data and to constrain measurements to the scanner's most accurate range, all setups were imported with a radial filter of 25~m. The TLS data were post-processed and aligned through a combination of manual and partial Iterative Closest Point (ICP) alignments within the Leica Register 360 software.

Our registration approach differs from previous methods by eliminating the need for physical reference targets, which were abandoned in CP2 due to previous impracticality. Instead, we developed a network-based registration protocol that meets RINGO goals:

\begin{enumerate}
\item Prioritizing high-density scans at close range ($< 12$~m) (uniform point density)
\item Employs selective manual intervention only when automated methods fail (accurate co-registration of scans)
\item leveraging VIO for initial alignment and improvisational setup planning (minimized occlusion and noise)
\end{enumerate}

During our registration process, we observed that when overlap between scans was limited, manual alignment often produced superior results compared to automated bundle optimization. Therefore, certain links (blue edges in Figure~\ref{fig:tls_network}) were designated as ``fixed'' to exclude them from bundle optimization, thereby avoiding automatic adjustments to visually verified alignments.

\subsubsection{Mobile Laser Scanning Protocol}

The backpack MLS configuration employs an oblique LiDAR angle for improved mid-high trunk capture, with cameras oriented away from the GNSS antenna to maximize upward canopy coverage. The elevated sensor position provides clearance advantages over ground-based systems for vegetation penetration.

Based on measurements from previous studies, traditional TLS methods can require several hours of field setup for comprehensive coverage of a single hectare. Our MLS approach allowed traversal of all 3 CPs in half a day. The operator followed paths to meet ICOS RINGO goals:

\begin{enumerate}
\item Maximize coverage of the plot area where understory traversability allowed it (uniform point density)
\item Ensure sufficient loop closures for pose graph optimization (accurate co-registration of scans)
\item Oblique LiDAR configuration to capture both trunks upward, and groundcover downward (minimized occlusion and noise)
\end{enumerate}

\subsection{Data Processing and Quality Control}

Multi-platform processing followed a systematic three-stage approach designed to preserve platform-specific characteristics while enabling cross-comparison. The workflow progressed from (i) inter-modality registration, to (ii) standardized ground classification, and (iii) height normalization and stratification analysis.

\subsubsection{Registration workflow}

All platforms were co-registered using a hierarchical transformation strategy anchored to the aerial RTK/PPK GNSS positioning, which provided the global geodetic reference frame~\cite{naesset1997estimating, naesset2002predicting}. TLS and MLS clouds were transformed to this frame using 4$\times$4 matrices included with the dataset. Final outputs are delivered in LAS format for compatibility with open-source forestry toolkits such as \texttt{lidR} in R~\cite{rasterlidR}. 

Semi-automatic inter-modality registration was adapted from hierarchical approaches developed for forest mapping~\cite{castorena_forestalign_2024}, with three successive steps: (i)~align the ground, (ii)~align stems and large branches, and (iii)~evaluate canopy and foliage alignment. Registration quality was assessed through cloud-to-cloud distance analysis (Figure~\ref{fig:multiplatform_integration}) and visual inspection of trunk correspondences. This workflow balances automation with manual checks, ensuring reproducibility while capturing local complexities in alpine forest structure.

\subsubsection{Ground classification and height normalization}

Ground points were classified using the Cloth Simulation Filter (CSF)~\cite{zhang2016csf} with standardized parameters (class threshold 0.5~m, cloth resolution 0.5~m, rigidness $2\,500$ iterations). Applying identical settings across modalities ensured a consistent ground-identification methodology, despite large differences in sampling density and geometry. 

\begin{figure}
\centering
\includegraphics[width=\textwidth]{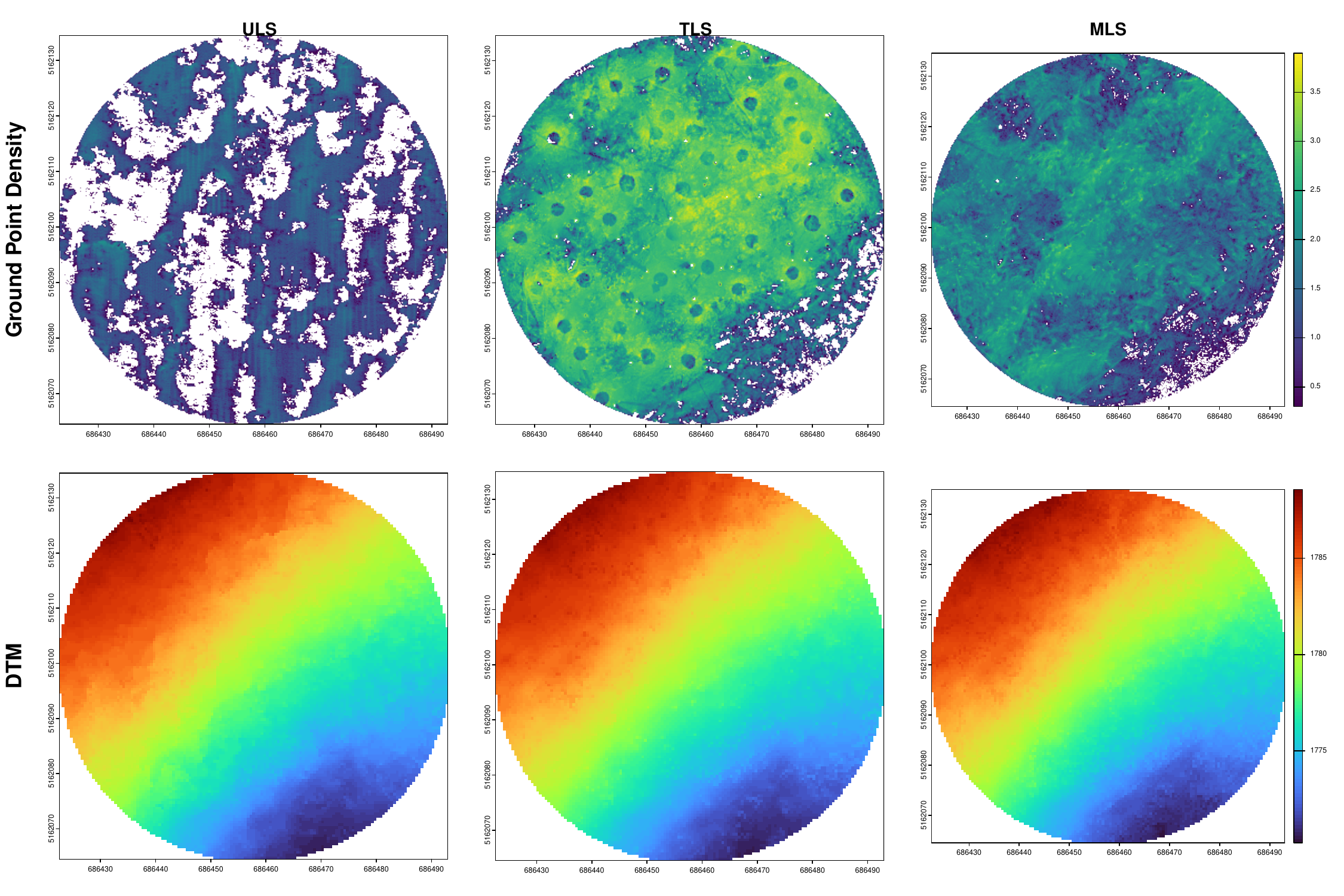}
\caption{Ground Points and resulting DTMs per platform. Top row visualizes ground point density at 0.2~m resolution in log$_{10}$ scale (range: 0.3-3.9). Bottom row presents corresponding DTMs at 0.5~m resolution (elevation range: 1770.2-1788.6~m) generated through CSF ground classification and kNN inverse distance weighting interpolation.}
\label{fig:ground_dtm_comparison}
\end{figure}

Ground point spatial distributions highlighted the contrasting sensing capabilities of ULS, TLS, and MLS: ULS provided broad canopy coverage with limited penetration, TLS achieved high near-ground densities but was affected by occlusion, and MLS captured continuous trajectories through accessible understory. Nevertheless, the resulting 0.5~m resolution digital terrain models (DTMs) were visually indistinguishable across modalities (Figure~\ref{fig:ground_dtm_comparison}), demonstrating the robustness of the CSF workflow and validating cross-platform geometric alignment essential for height normalization and subsequent stratification analysis~\cite{balestra_lidar_2024, muhojoki_benchmarking_2024}.

Height normalization removes ground elevations from all point clouds, enabling direct comparison of vegetation structure across platforms. The consistent elevation ranges achieved across platforms (-1.12 to -1.28~m minimum, 32.87 to 35.26~m maximum) validate successful cross-platform co-registration within the 35~m radius analysis area.

\subsubsection{Vertical distribution analysis and height stratification}

Following ground normalization, vegetation points were stratified into four vertical layers representing key structural components of the forest: understory (0–1.4~m), trunks (1.4–5~m), mid-canopy (5–13~m), and canopy tops ($ > 13$~m). This classification provides a framework for comparing how each platform samples different strata and for quantifying their complementary contributions to forest inventory. Stratification metrics are shown in Table~\ref{tab:stratification_metrics}.

\begin{figure}
\centering
\includegraphics[width=\textwidth]{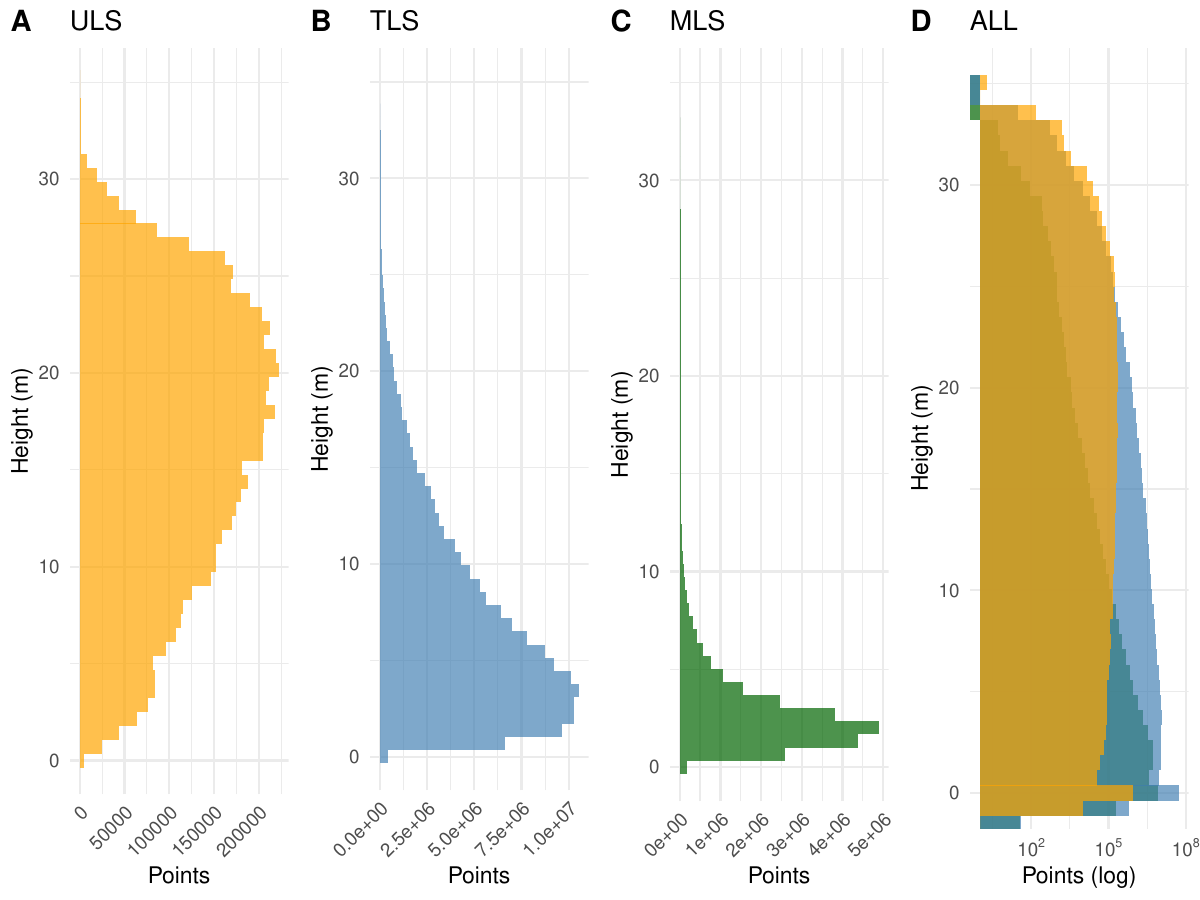}
\caption{Vegetation height distribution comparison across laser scanning platforms. (A-C) show vegetation-only points, normalized by the CSF DTM, and binned by Z-axis. ULS/ALS (A) is canopy-biased, TLS (B) is subcanopy-biased, but most point-dense, and MLS (C) is most height-constrained and subcanopy-biased. Combined log$_{10}$-scale comparison (D) enables cross-platform analysis despite order-of-magnitude differences in point density. Consistent height ranges across platforms (-1.5 to 35.5~m) indicate successful cross-platform registration, while platform-specific distribution shapes reflect fundamental differences in sensing geometry.}
\label{fig:height_distribution_comparison}
\end{figure}

Height distributions reveal distinct sensing biases (Figure~\ref{fig:height_distribution_comparison}). ULS is strongly canopy-oriented, TLS achieves the highest densities at trunk level but is affected by occlusion in dense understory, while MLS captures the understory and trunk strata efficiently, but provides limited penetration into upper canopy. Despite these differences, consistent vertical ranges across platforms (min $\approx$ –1.5~m; max $\approx$ 35.5~m) confirm successful cross-platform registration and support their integration for structural analysis~\cite{puliti_benchmarking_2024, wielgosz_segmentanytree_2024}.

\subsubsection{Spatial Point Density Mapping Across Height Strata}

\begin{figure}
\centering
\includegraphics[width=\textwidth]{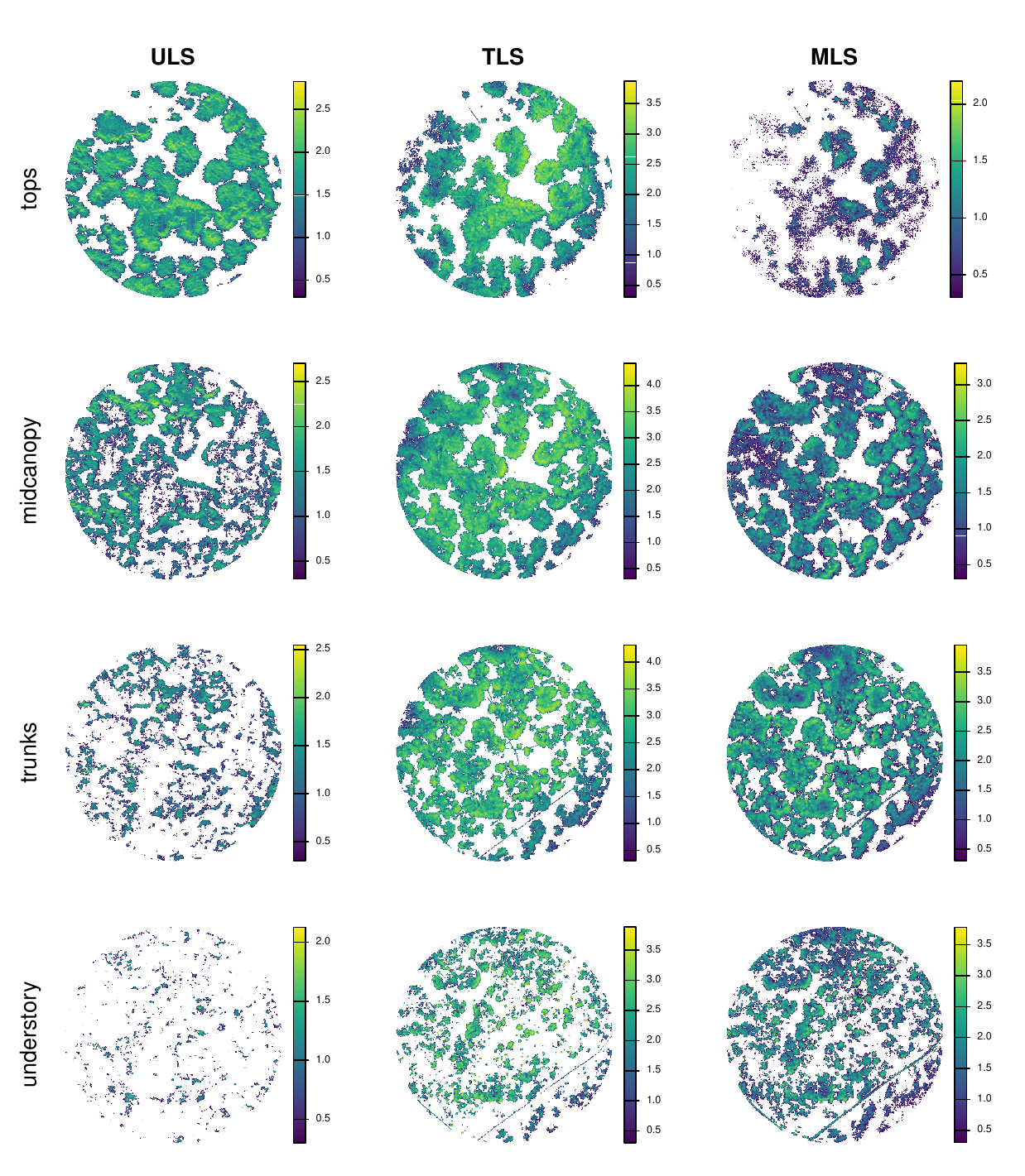}
\caption{Vertical distribution of point density across forest height strata for three scanning platforms, demonstrating complementary coverage characteristics essential for comprehensive forest inventory. Each circular plot represents point density on log$_{10}$ scale across height layers: understory (0-1.4~m), trunks (1.4-5~m), midcanopy (5-13~m), and canopy tops ($> 13$~m). ULS shows concentrated detection in upper canopy with limited ground penetration, TLS provides comprehensive coverage across all height strata with highest densities in trunk regions, while MLS exhibits strong understory and trunk detection with reduced canopy access.}
\label{fig:height_slice_comparison}
\end{figure}

To visualize coverage patterns, we generated 0.2~m resolution spatial density maps for each height stratum (Figure~\ref{fig:height_slice_comparison}). These maps illustrate how ULS provides uniform canopy coverage, TLS achieves detailed representation of trunks and near-ground vegetation, and MLS contributes dense sampling of understory and mid-trunk regions through operator-guided trajectories. 

By comparing spatial density distributions across strata, the dataset highlights platform-specific strengths and weaknesses, offering a practical reference for optimizing inventory protocols under ICOS and related monitoring frameworks. Such stratified mapping also provides a basis for evaluating segmentation methods, quantitative structure modeling, and allometric scaling approaches, and for benchmarking efficiency trade-offs between different scanning modalities~\cite{boucher2021sampling, su2021backpack, bauwens2016forest}, as well as evaluating segmentation and instance-level tree detection methods~\cite{puliti_for-instance_2023, wielgosz_point2treep2t_2023}.

\subsubsection{Cross-Platform Registration and Integration}

Multi-platform integration facilitated coordinate transformations over dense point-to-point alignment to preserve modality-specific characteristics for comparative analysis. The aerial survey (ULS) provided the absolute geodetic reference frame through PPK GNSS processing~\cite{naesset2002predicting}, TLS delivered local geometric fidelity via a registered scan network, and MLS accuracy was validated against both platforms using ground surfaces and stem bases as common reference features. Transformation matrices are provided with the dataset to ensure reproducibility and integration with alternative workflows.

Registration accuracy was evaluated through complementary approaches. Quantitative cloud-to-cloud distance analyses (Figure~\ref{fig:multiplatform_integration}) demonstrate alignment quality between MLS and ULS datasets, while visual inspection of stem alignments confirmed consistency across modalities. Reported metrics—including voxelized root-mean-square errors, bundle adjustment statistics, and overlap percentages—are included in the metadata files. These support independent validation and adaptation of the dataset for other registration frameworks~\cite{castorena_forestalign_2024}.

By combining automated registration with manual quality checks, the integration workflow ensures robustness under the challenging conditions of subalpine forest terrain, where steep slopes and dense understory often degrade alignment accuracy~\cite{ritter_towards_2020, muhojoki_benchmarking_2024}. The resulting co-registered dataset provides a reproducible benchmark for testing new methods in forest inventory, including SLAM- and VIO-based workflows~\cite{metzler2024vio, boche2024tightlycoupledlidarvisualinertialslamlargescale, oh2024evaluation}.

\section*{User Notes}

This dataset is intended as an open resource for the forest science community. It was developed with ICOS monitoring in mind, following the quality goals proposed in the RINGO project: consistent sampling across the plot, accurate co-registration between modalities, and reduced occlusion in complex stands. Marker-free protocols, systematic filtering, and loop-closure validation were applied to ensure robust integration of TLS, MLS, and ULS data under challenging alpine conditions, and supports the move toward operational “digital twins” for inventory and monitoring~\cite{holzinger_industry_2024}.

We invite users to explore the dataset in diverse ways. It provides opportunities to benchmark registration and segmentation algorithms, to compare field efficiency across scanning modalities, and to link structural attributes with above-ground biomass estimation through allometric or quantitative structure modelling approaches. Because TLS, MLS, and ULS each capture different parts of the canopy–understory continuum, the dataset is particularly useful for evaluating how modalities can complement each other in repeated inventories.  

At the same time, users should remain aware of site-specific challenges. On steep slopes, registration becomes less reliable, especially for MLS where walking paths are constrained. Dense understory benefits from MLS coverage but can introduce noise from multiple reflections. These limitations are not drawbacks, but rather reflect realistic field conditions in structurally complex ecosystems and highlight where different scanning approaches excel.

\section*{Author Contributions}
\begin{itemize}
    \item[ ] Conceptualization: MRC, AC, MC, and ET
    \item[ ] Data curation: MRC, AC
    \item[ ] Formal analysis: MRC, AC
    \item[ ] Funding acquisition: MC, ET, KVE
    \item[ ] Investigation: MRC, AC, ET, and MC
    \item[ ] Methodology: MRC, MC
    \item[ ] Software: MRC
    \item[ ] Supervision: MC, ET, and KVE
    \item[ ] Validation: MC, ET, and KVE
    \item[ ] Writing - original draft: MRC
    \item[ ] Writing - review \& editing: MRC, MC, ET, AC, and KVE   
\end{itemize}

\section*{Competing Interests}
The authors declare there are no competing interests.
\section*{Data Availability}
Data supporting this article, with the exclusion of raw data, are available at \cite{dataset} through the following DOI: \texttt{10.5281/zenodo.17186174}. The dataset includes point clouds, registration data, and comprehensive metadata.  All data are provided under Creative Commons Attribution 4.0 International License. Raw sensor data (i.e., rosbags) are available upon request.

\section*{Acknowledgement}
The authors thank the ICOS Principal Investigators for providing facilities during autumn 2024 data collection campaigns. We acknowledge Leonardo Montagnani of the ICOS IT-Ren Selva-Verde station for administering the flux tower site since 1997 and facilitating site access, internet connectivity, and laboratory facilities during November field operations.

Special recognition goes to Professor Maurice Fallon of the University of Oxford and the Digiforest consortium (EU Horizon, Grant ID: 101070405) for providing the Frontier mobile mapping payload that enabled comprehensive mobile laser scanning data collection.

AC was partially funded by FORMA (EFRE/FESR 2021-2027). Project id: EFRE1079.

ET was partially funded by INEST (PNRR - Italian National Plan for Recovery and Resilience),
Project id: ECS00000043.

We thank the technical and field support teams who assisted with equipment deployment, data collection logistics, and site coordination throughout the multi-platform scanning campaigns.

\section{Abbreviations}{
The following abbreviations are used in this manuscript:

\noindent 
\begin{tabular}{@{}ll}
ALS & Aerial Laser Scanning\\
CSF & Cloth Simulation Filter\\
DTM & Digital Terrain Model\\
GNSS & Global Navigation Satellite System\\
ICOS & Integrated Carbon Observation System\\
IMU & Inertial Measurement Unit\\
LAS & LiDAR Data Exchange Format\\
LIO & LiDAR-Inertial Odometry\\
LiDAR & Light Detection and Ranging\\
MLS & Mobile Laser Scanning\\
PPK & Post-Processed Kinematic\\
SLAM & Simultaneous Localization and Mapping\\
TLS & Terrestrial Laser Scanning\\
ULS & Unmanned Laser Scanning\\
VIO & Visual-Inertial Odometry
\end{tabular}
}

\bibliographystyle{agsm}
\bibliography{data2025}

\end{document}